\documentclass[journal,9pt]{IEEEtran}

\usepackage[cmex10]{amsmath}
\usepackage{epsfig}
\usepackage{amsthm}
\usepackage{url}
\usepackage{multirow}
\usepackage{rotating}
\usepackage{amssymb}
\usepackage{graphicx,amsfonts}
\usepackage{algorithm,algorithmic}
\usepackage{color}

\usepackage{cite}
\ifCLASSINFOpdf
\else
\fi

\usepackage{stfloats}

\newtheorem{prop}{Proposition}

\newtheorem{rmk}{Remark}

\newcommand{\cH}{\mathcal{H}}

\newcommand{\cL}{\mathcal{L}}

\newcommand{\bU}{\boldsymbol{U}}

\newcommand{\bT}{\boldsymbol{T}}

\newcommand{\bPhi}{\boldsymbol{\Phi}}

\newcommand{\bx}{\boldsymbol{x}}

\newcommand{\by}{\boldsymbol{y}}
\newcommand{\bu}{\boldsymbol{u}}
\newcommand{\bv}{\boldsymbol{v}}
\newcommand{\bz}{\boldsymbol{z}}

\newcommand{\bc}{\boldsymbol{c}}

\newcommand{\bw}{\boldsymbol{w}}

\newcommand{\bZero}{\boldsymbol{0}}

\newcommand{\beq}{\begin{equation}}
\newcommand{\eeq}{\end{equation}}
\newcommand{\beqn}{\begin{eqnarray}}
\newcommand{\eeqn}{\end{eqnarray}}
\newcommand{\beqns}{\begin{eqnarray*}}
\newcommand{\eeqns}{\end{eqnarray*}}

\newcommand{\R}{\mathbb{R}}
\newcommand{\HH}{\mathbb{H}}

\newcommand{\C}{\mathbb{C}}

\newcommand{\F}{\mathbb{F}}
\newcommand{\N}{\mathbb{N}}

\newcommand{\pend}{\hfill$\square$\\}

\makeatletter
\newcommand{\bdiv}{\mathop{\operator@font div}}
\makeatother

\makeatletter
\newcommand{\diag}{\mathop{\operator@font diag}}
\makeatother

\makeatletter
\newcommand{\conv}{\mathop{\operator@font conv}}
\makeatother

\makeatletter
\newcommand{\sign}{\mathop{\operator@font sign}}
\makeatother \makeatletter

\makeatletter
\newcommand{\proj}{\mathop{\operator@font proj}}
\makeatother \makeatletter

\makeatletter
\newcommand{\spa}{\mathop{\operator@font span}}
\makeatother \makeatletter

\makeatletter
\newcommand{\epi}{\mathop{\operator@font epi}}
\makeatother \makeatletter

\makeatletter
\newcommand{\dom}{\mathop{\operator@font dom}}
\makeatother \makeatletter

\begin{document}

% paper title
% can use linebreaks \\ within to get better formatting as desired
\title{The Augmented Complex Kernel LMS}
%
%
% author names and IEEE memberships
% note positions of commas and nonbreaking spaces ( ~ ) LaTeX will not break
% a structure at a ~ so this keeps an author's name from being broken across
% two lines.
% use \thanks{} to gain access to the first footnote area
% a separate \thanks must be used for each paragraph as LaTeX2e's \thanks
% was not built to handle multiple paragraphs
%

\author{Pantelis Bouboulis,~\IEEEmembership{Member,~IEEE,}
        and~Sergios Theodoridis,~\IEEEmembership{Fellow,~IEEE}
        and~Michael Mavroforakis% <-this % stops a space
\thanks{Copyright (c) 2010 IEEE. Personal use of this material is permitted.
However, permission to use this material for any other purposes must be obtained from the
IEEE by sending a request to pubs-permissions@ieee.org.}
\thanks{P. Bouboulis is with the Department
of Informatics and Telecommunications, University of Athens, Greece,
e-mail: bouboulis@di.uoa.gr.}% <-this % stops a space
%\thanks{Manuscript received ....}
\thanks{S. Theodoridis is with the Department
of Informatics and Telecommunications, University of Athens, Greece,
and the Research Academic Computer Technology Institute, Patra, Greece.
e-mail: stheodor@di.uoa.gr.}
\thanks{M. Mavroforakis is with the computational Biomedicine Lab, Department
of Computer Science, University of Houston, Texas,
e-mail: mmavrof@uh.edu.}
}
% note the % following the last \IEEEmembership and also \thanks -
% these prevent an unwanted space from occurring between the last author name
% and the end of the author line. i.e., if you had this:
%
% \author{....lastname \thanks{...} \thanks{...} }
%                     ^------------^------------^----Do not want these spaces!
%
% a space would be appended to the last name and could cause every name on that
% line to be shifted left slightly. This is one of those "LaTeX things". For
% instance, "\textbf{A} \textbf{B}" will typeset as "A B" not "AB". To get
% "AB" then you have to do: "\textbf{A}\textbf{B}"
% \thanks is no different in this regard, so shield the last } of each \thanks
% that ends a line with a % and do not let a space in before the next \thanks.
% Spaces after \IEEEmembership other than the last one are OK (and needed) as
% you are supposed to have spaces between the names. For what it is worth,
% this is a minor point as most people would not even notice if the said evil
% space somehow managed to creep in.

% The paper headers
\markboth{IEEE Transactions on Signal Processing}%
{Bouboulis \MakeLowercase{\textit{et al.}}: The Augmented Complex Kernel LMS}
% The only time the second header will appear is for the odd numbered pages
% after the title page when using the twoside option.
%
% *** Note that you probably will NOT want to include the author's ***
% *** name in the headers of peer review papers.                   ***
% You can use \ifCLASSOPTIONpeerreview for conditional compilation here if
% you desire.

% If you want to put a publisher's ID mark on the page you can do it like
% this:
%\IEEEpubid{0000--0000/00\$00.00~\copyright~2007 IEEE}
% Remember, if you use this you must call \IEEEpubidadjcol in the second
% column for its text to clear the IEEEpubid mark.

% use for special paper notices
%\IEEEspecialpapernotice{(Invited Paper)}

% make the title area
\maketitle

\begin{abstract}
Recently, a unified framework for adaptive kernel based signal processing of complex data was presented by the authors, which, besides offering techniques to map the input data to complex Reproducing Kernel Hilbert Spaces, developed a suitable Wirtinger-like Calculus for general  Hilbert Spaces. In this short paper, the extended Wirtinger's calculus is adopted to derive complex kernel-based widely-linear estimation filters. Furthermore, we illuminate several important characteristics of the widely linear filters. We show that, although in many cases the gains from adopting widely linear estimation filters, as alternatives to ordinary linear ones, are rudimentary, for the case of kernel based widely linear filters  significant performance improvements can be obtained.
\end{abstract}
% IEEEtran.cls defaults to using nonbold math in the Abstract.
% This preserves the distinction between vectors and scalars. However,
% if the journal you are submitting to favors bold math in the abstract,
% then you can use LaTeX's standard command \boldmath at the very start
% of the abstract to achieve this. Many IEEE journals frown on math
% in the abstract anyway.

% Note that keywords are not normally used for peerreview papers.
%\begin{IEEEkeywords}
%Complex RKHS, Wirtinger's Calculus, KLMS, Adaptive filter, complex signal processing
%\end{IEEEkeywords}

% For peer review papers, you can put extra information on the cover
% page as needed:
% \ifCLASSOPTIONpeerreview
% \begin{center} \bfseries EDICS Category: 3-BBND \end{center}
% \fi
%
% For peerreview papers, this IEEEtran command inserts a page break and
% creates the second title. It will be ignored for other modes.
\IEEEpeerreviewmaketitle

%% main text

%--------------------------------------------------------------------------------
\section{Introduction}\label{SEC:INTRO}
%--------------------------------------------------------------------------------
Kernel-based processing is gaining in popularity within the Signal Processing community \cite{Liu_2010_10644, Xu_2008_9259, Kivinen_2004_11230, Engel_2004_11231, Slavakis_2008_9257, Slavakis_2009_10648, Bouboulis_2010_10642}, as it provides an efficient toolbox for treating non-linear problems. The main advantage of this procedure is that it transforms the original  non-linear task, in a low dimensional space, into a linear one, that is performed in a higher dimensionality (possible infinite Hilbert) space $\cH$. This is equivalent with solving a nonlinear problem in the original space.
The space $\cH$ is implicitly chosen via a kernel function that defines the associated inner product.

However, most of the kernel-based techniques were designed to process real data. Until recently, no kernel-based methodology for treating complex signals had been developed, in spite of their potential interest in a number of applications. In \cite{Bouboulis_2011_10643}, a framework based on complex RKHS was presented to solve this problem. Its main contributions are:  a) the development of a wide framework that allows  real-valued kernel algorithms to be extended to treat complex data efficiently, taking advantage of a technique called \textit{complexification} of real RKHSs, b) the elevation from obscurity of the pure complex kernels (such as the complex Gaussian one) as a tool for kernel based adaptive processing of complex signals and c) the extension of \textit{Wirtinger's Calculus} in complex RKHSs as a means for an elegant and efficient computation of the gradients, which are involved in the derivation of adaptive learning algorithms.

Complex-valued signals arise frequently in applications as diverse as communications, biomedicine, radar, etc. The complex domain not only provides a convenient and elegant representation for these signals, but also a natural way to preserve their characteristics and to handle transformations that need to be performed. In the more traditional setting, treating complex signals is often followed by (implicitly) assuming the \emph{circularity} of the signal. Circularity is intimately related to the rotation in the geometric sense. A complex random variable $Z$ is called circular, if for any angle $\phi$ both $Z$ and $Z e^{i\phi}$ (i.e., the rotation of $Z$ by angle $\phi$) follow the same probability distribution \cite{Mandic_2009_10646, Adali_2010_10640}. Naturally, this assumption limits the area for applications, since many practical signals exhibit non-circular characteristics. Thus, following the ideas originated by Picinbono and Chevalier in \cite{Picinbono_1995_10647, Picinbono_1994_10659}, on-going research is focusing on the \emph{widely linear} filters (or \emph{augmented} filters) in the complex domain (see for example \cite{Mandic_2009_10646, Adali_2010_10640, Novey_2008_10638, Cacciapuoti_2008_10667, Kuchi_2009_10669, Mattera_2003_10672, Pun_2008_10674, Xia_2010_11626, Aghaei_2010_10665, NavarroMoreno_2009_10670, Kuchi_2009_10669}). The main characteristic of such filters is that they exploit simultaneously both the original signal as well as its conjugate analogue. An important characteristic of widely linear estimation is that it captures the full second-order statistics of a given complex signal, especially in the case where the signal is non-circular, by considering both the covariance and the pseudo-covariance matrices \cite{Picinbono_1995_10647, Picinbono_1997_10660}. Thus, for a general complex signal, the optimal linear filter (i.e., widely linear) is linear both in $z$ and $z^*$.

The main contribution of the present paper is twofold. First, we explore the main concepts of widely linear estimation from a new perspective, showing why widely linear estimation can potentially lead to improved performance compared to complex linear estimation. Moreover, we employ the framework of \cite{Bouboulis_2011_10643} to develop widely linear adaptive filters in complex RKHS, to solve nonlinear filtering tasks. Thus, we introduce, for the first time, the notion of widely linear filters in infinite dimensional Hilbert spaces. Our findings indicate that the ``natural'' choice for kernels, in the context of the widely linear filtering structure, are the pure complex kernels.  In contrast, combining the widely linear structure with kernels that result from complexification of real kernels, does not enhance performance, compared to that obtained with the complexified kernel-based linear filters\footnote{This is because complexification implicitly adds a conjugate component to the adopted model.}.

\textcolor{red}{
The problem of widely linear estimation using kernels has been also considered in \cite{kuh09,kuh10}. Although in these papers the authors claim that they solve the problem of augmented complex signal processing using kernels, a careful study reveals a lot of points that are not clearly described (possibly because they are addressed to a conference). Moreover (and this is more important) there are several significant inconsistencies with the traditional theory of Reproducing Kernel Hilbert spaces (RKHS), which is the main mathematical tool of any kernel based method. For example, in both papers the complex data  are mapped to the so called feature space using a mapping $\bPhi: C^n \rightarrow \C^d$, instead of the traditional approach, which considers a feature map of the form $\bPhi:\C^n\rightarrow\HH$, $\bPhi(\bz)=\kappa(\cdot,\bz)$, where $\HH$ is the RKHS induced by the kernel $\kappa$. It is evident, that under this context, $\bPhi$  maps the data to an Euclidean space. Following this mapping of the data, the presented methods build some kernel matrices of the form  $(\bPhi(\bx_1), \dots, \bPhi(\bx_M))$. However, the case of treating the problem in finite dimensional Euclidean spaces, is a rather trivial extension of previously known results. Moreover, in those papers, the notion of fully complex kernels is neither mentioned and even more treated. However, as we prove in the present paper, this is the case where the augmented structure has a meaning, in the sense that can offer advantages. Using real kernels, under the complexification theory, in the context of augmented filters, offers no advantages.
}

The paper is organized as follows. We start with a brief introduction to complex RKHSs  and Wirtinger's Calculus in Section \ref{SEC:PRELIM}.  In Section \ref{SEC:Widely_Linear}, we describe the concept of widely linear estimation and show why this is better than complex linear estimation. Finally, widely linear kernel based adaptive filters are described in section \ref{SEC:ACKLMS}. Experiments are provided in section \ref{SEC:EXPER}. Section \ref{SEC:CONCL} concludes the paper.

%--------------------------------------------------------------------------------
\section{Preliminaries}\label{SEC:PRELIM}
%--------------------------------------------------------------------------------
Throughout the paper, we will denote the set of all integers, real and complex numbers by $\N$, $\R$ and $\C$ respectively. Vector or matrix valued quantities appear in boldfaced symbols. A RKHS \cite{Aronszajn_1950_9268} is a Hilbert space $\cH$ over a field $\F$ for which there exists a positive definite function $\kappa:X\times X\rightarrow\F$ with the following two important properties: a) For every $x\in X$, $\kappa(\cdot,x)$ belongs to $\cH$ and b) $\kappa$ has the so called \textit{reproducing property}, i.e.,
$f(x)=\langle f,\kappa(\cdot, x)\rangle_\cH, \textrm{ for all } f\in\cH$,
in particular $\kappa(x,y)=\langle \kappa(\cdot, y), \kappa(\cdot, x)\rangle_\cH$.
The map $\Phi:X\rightarrow\cH:\Phi(x)=\kappa(\cdot,x)$ is called the \textit{feature map} of $\cH$.

Although the underlying theory has been developed by the mathematicians for  general complex reproducing kernels and their associated RKHSs, mostly the real kernels have been considered by the machine learning and signal processing communities \cite{Scholkopf_2002_2276, Bouboulis_2011_11457}. Some of the most widely used kernel in the literature are the \textit{Gaussian RBF}, i.e., $\kappa_{\sigma,\R^d}(\bx,\by) : = \exp\left(-\frac{\sum_{k=1}^{d}(x_k-y_k)^2}{\sigma^2}\right)$,
defined for $\bx, \by \in \R^d$, where $\sigma$ is a free positive parameter  and the \textit{polynomial kernel}: $\kappa_d(\bx,\by) : = \left(1 + \bx^T \by\right)^d$, for $d\in\N$. Many more can be found in \cite{Scholkopf_2002_2276, Theodoridis_2008_9253}. Complex reproducing kernels, that have been extensively studied by the mathematicians, are, among others, the \textit{Szego kernels} and the \textit{Bergman kernels}. Another important complex kernel is the \textit{complex Gaussian kernel}, which is defined as:
$\kappa_{\sigma,\C^d}(\bz,\bw) : = \exp\left(-\frac{\sum_{k=1}^{d}(z_k-w_k^*)^2}{\sigma^2}\right)$,
where $\bz,\bw\in\C^d$, $z_k$ denotes the $k$-th component of the complex vector $\bz\in\C^d$ and $\exp(\cdot)$ is the extended exponential function in the complex domain.

To generate kernel adaptive filtering algorithms on complex domains, according to \cite{Bouboulis_2011_10643}, one can adopt two methodologies. A first straightforward approach is to use directly a complex RKHS, using one of the complex kernels given in section \ref{SEC:PRELIM} and map the original data to the complex RKHS through the associated feature map $\bPhi(x)=\kappa(\cdot,x)$. Another alternative technique is to use real kernels through a rationale that is called \textit{complexification} of real RKHSs. This method has the advantage of allowing  modeling in complex RKHSs using popular well-established and well understood, from a performance point of view, real kernels (e.g., gaussian, polynomial, e.t.c.). In the first case, we map the data directly to the complex RKHS, using the corresponding complex feature map $\bPhi(\bz) = \kappa_{\C}(\cdot,\bz)$, while in the complexification scenario we employ the map $\hat\bPhi(\bz) = \Phi(\bx,\by) + i\Phi(\bx,\by)$, where $\bz=\bx+i\by$, and $\Phi$ is the feature map of the chosen real kernel $\kappa_{\R}$, i.e., $\Phi(\bx, \by) = \kappa_{\R}(\cdot, (\bx, \by))$.

In order to compute the gradients of real valued cost functions, that are defined on complex domains, we adopt the rationale of Wirtinger's calculus \cite{Wirtinger_1927_10651}. This was brought into light recently \cite{Mandic_2009_10646, Adali_2010_10640, Picinbono_1995_10647, Novey_2008_10638, Li_2008_10664}, as a means to compute, in an efficient and elegant way,  gradients of real valued cost functions that are defined on complex domains ($\C^\nu$). It is based on simple rules and principles, which bear a great resemblance to the rules of the standard complex derivative, and it greatly simplifies the calculations of the respective derivatives. The difficulty with real valued cost functions is that they do not obey the Cauchy-Riemann conditions and are not differentiable in the complex domain. The alternative to Wirtinger's calculus would be  to consider the complex variables as pairs of two real ones and employ the common real partial derivatives. However, this approach, usually, is more time consuming and leads to more cumbersome expressions. In \cite{Bouboulis_2011_10643}, the notion of Wirtinger's calculus was extended to general complex Hilbert spaces, providing the tool to compute the gradients that are needed to develop kernel-based algorithms for treating complex data.
In the same paper, the aforementioned toolbox was employed in the context of the complex LMS and two realizations of the complex kernel LMS algorithm were developed. The first one, which is denoted as NCKLMS1 adopts the complexification procedure and the second one, which is denoted as NCKLMS2, uses the complex gaussian kernel\footnote{Due to the fact that there are a lot of algorithms mentioned in the paper, Table \ref{TAB:Acro} presents their acronyms for the reader's conveniance.}.

\section{Widely linear estimation filters}\label{SEC:Widely_Linear}
As mentioned in the introduction, ongoing research in complex signal processing is mainly focused on the so called \textit{widely linear} estimation filters, or \textit{augmented} filters, as they are also known. These are filters that take into account both the original values of the signal data and their conjugates. For example, in a typical LMS task, we estimate the output as $\hat d(n)=\bw^H \bz$ and the step update as $\bw(n)=\bw(n-1) + \mu e^*(n)\bz(n)$. In this case, $\hat d(n)$ is a linear estimation filter. However, the linearity property is taken with respect to the field of complex numbers. Picinbono and Chevalier, in \cite{Picinbono_1995_10647}, proposed an alternative approach. They estimated the filter's output as $\tilde d(n) = \bw^H \bz + \bv^H \bz^*$ and showed that it provides better results in terms of the mean square error. This, of course, is expected since $\tilde d(n)$ provides a more rich representation than $\hat d(n)$. On the other hand, $\tilde d(n)$ is no longer linear over the field $\C$. It is linear, however, over the real numbers $\R$. To emphasize this difference, in the relative literature, $\hat d(n)$ is often called $\C$-linear, while $\tilde d(n)$ is called $\R-linear$.

In \cite{Picinbono_1995_10647, Picinbono_1997_10660, Schreier_2003_11158}, it is shown that the widely linear estimation filter is able to capture the second order statistical characteristics of the signal which are essential, for non-circular sources. Although in many relative works this is highlighted as the main reason for adopting widely linear techniques, in this paper, we will highlight a different perspective.

Our starting point will be the definition of linearity. Prior to it, it should be clarified, that complex processing is equivalent with processing two real signals in the respective Euclidean (Hilbert) spaces. The advantage of using complex algebra is that the algorithm and/or the solution may be described in a more compact form. Moreover, the complex algebra allows for a more intuitive understanding of the problem, as many geometric transformations can be easily described using complex algebra in an elegant way. Finally, the application of Wirtinger's calculus greatly simplifies the calculations needed for the gradients of real valued cost functions.

Having this in mind, we now turn our attention to a typical complex LMS task. Let $\bz(n)\in\C^{\nu}$ and $d(n)\in\C$ be the input and the output of the original filter. We estimate the output of the filter using a $\C$-linear response $\hat d(n) = \bw^H\bz(n)$. The typical complex LMS task aims to compute $\bw\in\C^{\nu}$, such that the error $E[|d(n) - \hat d(n)|^2]$ is minimized. If we set $\bw = \bw_r + i\bw_i$ and $\bz=\bx + i\by$, we take that
\begin{align}\label{EQ:C_lin1}
\hat d(n) = \bw_r^T\bx + \bw_i^T\by + i(\bw_r^T\by - \bw_i^T\bx).
\end{align}

However, the real essence behind a complex filter operation is the following: Given two real vectors, $\bx(n)$ and $\by(n)$, compute linear filters in order to estimate two real processes, $d_r(n)$ and $d_i(n)$, in an optimal way, that jointly cares for both $d_r(n)$ and $d_i(n)$. Let us express the problem in its multichannel formulation, using real variables only, i.e.,
\begin{align*}
\left(\begin{matrix}\bar d_r(n) \cr \bar d_i(n)\end{matrix}\right) = \left(\begin{matrix}\bu_{1,1}^T & \bu_{1,2}^T \cr \bu_{2,1}^T & \bu_{2,2}^T\end{matrix}\right)\cdot\left(\begin{matrix}\bx \cr \by\end{matrix}\right) \equiv \bU \cdot\left(\begin{matrix}\bx \cr \by\end{matrix}\right)
\end{align*}

From a mathematical point of view, this is the definition of a linear operator from $\R^{2\nu}\rightarrow\R^2$. The elements of $\bU$ are computed
such that both $|d_r(n) - \bar d_r(n)|^2$ and $|d_i(n) - \bar d_i(n)|^2$ are jointly minimized. This leads to the so called Dual Real Channel (DRC) formulation. Thus, we take the relations
$\bar d_r(n) = \bu_{1,1}^T\bx + \bu_{1,2}^T\by$ and
$\bar d_i(n) = \bu_{2,1}^T\bx + \bu_{2,2}^T\by$.
In complex notation, if we consider that $\bar d(n) = \bar d_r(n) + i\bar d_i(n)$, we obtain the expression
\begin{align}\label{EQ:DRC1}
\bar d(n) = \bu_{1,1}^T\bx + \bu_{1,2}^T\by + i\left( \bu_{2,1}^T\bx + \bu_{2,2}^T\by\right).
\end{align}
It is easy to see that the DRC approach expressed by relation (\ref{EQ:DRC1}) adopts a richer representation than that of the traditional LMS in (\ref{EQ:C_lin1}). Moreover, it takes a few lines of elementary algebra to show that an equivalent expression of (\ref{EQ:DRC1}) is the widely linear estimation filter, i.e., $\bar d(n) = \tilde d(n)$. This will be our kick off point to define the task in a general Hilbert space. In general, we can prove the following:

\begin{prop}\label{PROP:main_widely}
Consider a real\footnote{By the term real (complex) Hilbert space, we mean a Hilbert space over the field of real numbers $\R$ (complex numbers $\C$). } Hilbert space $\cH$, the real Hilbert space $\cH^2$ and the complex Hilbert space $\HH=\cH+i\cH$. Then any continuous linear function $\bT:\cH^2\rightarrow\R^2$ can be expressed in complex notation as
\begin{align}\label{EQ:w_lin}
\bT(\bx,\by) = \bT(\bx+i\by) = \bT(\bz) = \left\langle \bz, \bw \right\rangle_{\HH} + \left\langle \bz^*, \bv \right\rangle_{\HH},
\end{align}
for some $\bw, \bv\in\HH$, where $\langle\cdot,\cdot\rangle_{\HH}$ is the respective inner product of $\HH$.
\end{prop}

\proof
Recall that any element of $\cH^2$ or $\R^2$ can be expressed either as a vector or as a complex element (complex notation). Any function $\bT:\cH^2\rightarrow\R^2$ takes the form $\bT(\bx,\by)=\left(T_1(\bx,\by), T_2(\bx,\by)\right)^T$. As $\bT$ is linear, both $T_1$, $T_2$ are also linear. Thus, as a consequence of the Riesz's representation theorem, there is $\bu_1=(\bu_{1,1}^T, \bu_{1,2}^T)^T\in\cH^2$, such that $T_1(\bx,\by)=\langle(\bx,\by)^T,(\bu_{1,1}^T, \bu_{1,2}^T)^T\rangle_{\cH^2}$. Similarly, $T_2(\bx,\by)=\langle(\bx,\by)^T,(\bu_{2,1}^T, \bu_{2,2}^T)^T\rangle_{\cH^2}$, for some $\bu_2=(\bu_{2,1}^T, \bu_{2,2}^T)^T\in\cH^2$.

Thus, in complex notation, $\bT$ can be written as
\begin{align}
\bT(\bx,\by) &= T_1(\bx,\by) + iT_2(\bx,\by)\nonumber\\
&= \langle\bx,\bu_{1,1}\rangle_{\cH} + \langle\by,\bu_{1,2}\rangle_{\cH} + i(\langle\bx,\bu_{2,1}\rangle_{\cH} + \langle\by,\bu_{2,2}\rangle_{\cH}). \label{EQ:T_lin}
\end{align}
We define $\bw_1, \bw_2, \bv_1, \bv_2\in\cH$ as follows:
$\bw_1 = \frac{\bu_{1,1} + \bu_{2,2}}{2},  \quad \bw_2 = \frac{\bu_{1,2} - \bu_{2,1}}{2},
\bv_1 = \frac{\bu_{1,1} - \bu_{2,2}}{2},  \quad \bv_2 = - \frac{\bu_{2,1} + \bu_{1,2}}{2}$.
Then, $\bu_{1,1} = \bw_1 + \bv_1$, $\bu_{1,2}=\bw_2 - \bv_2$, $\bu_{2,2}=\bw_1 - \bv_1$ and $\bu_{2,1}=-\bw_2-\bv_2$.
Substituting in (\ref{EQ:T_lin}) we take:
\begin{align*}
\bT(\bz) =& \bT(\bx,\by) = \langle\bx,\bw_1\rangle_{\cH} + \langle\bx,\bv_1\rangle_{\cH} + \langle\by,\bw_2\rangle_{\cH} - \langle\by,\bv_2\rangle_{\cH}\\
 &+ i\left( -\langle\bx,\bw_2\rangle_{\cH} - \langle\bx,\bv_2\rangle_{\cH} + \langle\by,\bw_1\rangle_{\cH} - \langle\by,\bv_1\rangle_{\cH}\right)\\
 =& \langle \bx+i\by, \bw_1+i\bw_2\rangle_{\HH} + \langle \bx-i\by, \bv_1+i\bv_2\rangle_{\HH}\\
=& \langle \bz, \bw\rangle_{\HH} + \langle \bz^*, \bv\rangle_{\HH},
\end{align*}
where $\bz=\bx+i\by$, $\bw = \bw_1+i\bw_2$, $\bv = \bv_1+i\bv_2$. \pend

\begin{rmk}\label{REM:rem1}
In view of proposition \ref{PROP:main_widely}, one understands that the original formulation of the complex LMS was rather ``unorthodox", as it excludes a large class of linear functions from being considered in the estimation process. It is evident, that the linearity with respect to the field of complex numbers is restricted, compared to the linearity that underlies the DRC approach, which is more natural. Thus, the correct complex linear estimation is $\bT(\bz) = \left\langle \bz, \bw \right\rangle_{\HH} + \left\langle \bz^*, \bv \right\rangle_{\HH}$ rather than $\bT(\bz) = \left\langle \bz, \bw \right\rangle_{\HH}$.
\end{rmk}

\begin{rmk}\label{REM:rem2}
Any $\R$-linear function can be expressed as in (\ref{EQ:w_lin}).
\end{rmk}

\begin{rmk}
Let $e(n) = d(n) - \tilde d(n)$, be the error of the widely linear estimation, where $d(n)$ is the desired response. If we assume that the error follows the complex normal distribution of van den Boss \cite{Boss_1995_11247}, with covariance matrix
$V = \left( \begin{matrix} 1 &  0 \cr 0  &  1 \end{matrix}\right)$,
we can easily derive that the maximum likelihood estimator, for $\bw,\bv$ is equivalent with minimizing the square error, as it is the case in widely linear LMS.
\end{rmk}

\section{Widely Linear estimation in complex RKHS}\label{SEC:ACKLMS}
In this section, we will develop realizations of the \textit{Augmented Complex Kernel LMS} (ACKLMS) algorithm using either pure complex kernels, or real kernels under the complexification trick. We show that ACKLMS offers substantial improvements versus complex kernel LMS (CKLMS), when the complex gaussian kernel is employed. On the other hand, the ACKLMS, which is developed under the complexification trick, degenerates to the standard CKLMS.

Consider the sequence of examples $(\bz(1),d(1))$, $(\bz(2),d(2))$, $\dots$, $(\bz(N),d(N))$, where $d(n)\in\C$, $\bz(n)\in\C^\nu$, $\bz(n)=\bx(n) + i \by(n)$, $\bx(n), \by(n)\in\R^\nu$, for $n=1,\dots,N$. Consider, also, a real RKHS $\cH$, the real Hilbert space $\cH^2$ and the complex Hilbert space $\HH=\cH + i\cH$. We model the widely linear estimation filter in $\HH$ as:
\begin{align}
\tilde d(n) &= \left\langle \bPhi(\bz(n)), \bw(n-1)\right\rangle_{\HH} + \left\langle \bPhi^*(\bz(n)), \bv(n-1)\right\rangle_{\HH},\label{EQ:estim1}
\end{align}
where $\bPhi$ is an appropriate function that maps the input data to the feature space $\HH$. This is equivalent with transforming the data to a complex RKHS and applying a widely linear complex LMS to the transformed data. The objective of the ACKLMS is to estimate, $\bw$ and $\bv$, so that to minimize $E\left[\cL_n(\bw)\right]$, where
$\cL_n(\bw) =  |e(n)|^2 = \left|d(n) - \tilde d(n)\right|^2$,
at each time instance $n$.

Applying the rules of Wirtinger's calculus in complex RKHS, we can easily deduce that $\frac{\partial \cL(n)}{\partial \bw^*}= -\bPhi(\bz(n))\cdot e^*(n)$, $\frac{\partial \cL(n)}{\partial \bv^*}= -\bPhi^*(\bz(n))\cdot e^*(n)$. Thus, the step updates of the ACKLMS are
$\bw(n) = \bw(n-1) + \mu \bPhi(\bz(n)) e^*(n)$,
$\bv(n) = \bv(n-1) + \mu \bPhi^*(\bz(n)) e^*(n)$.
Assuming that $\bw(0)=\bv(0)=\bZero$, the repeated application of the weight-update equations gives:
\begin{align}
\bw(n-1) = \mu \sum_{k=0}^{n-1}\bPhi(\bz(k)) e^*(k), \label{EQ:final_w}\\
\bv(n-1) = \mu \sum_{k=0}^{n-1}\bPhi^*(\bz(k)) e^*(k).\label{EQ:final_v}
\end{align}
Combining (\ref{EQ:estim1}), (\ref{EQ:final_w}) and (\ref{EQ:final_v})  leads us to conclude that the filter's output, at iteration $n$, becomes:
\begin{align*}
\tilde d(n) =& \mu \sum_{k=1}^{n-1} e(k)\left\langle\bPhi(\bz(n)), \bPhi(\bz(k))\right\rangle_{\HH} \\
&+   \mu \sum_{k=1}^{n-1} e(k) \left\langle\bPhi^*(\bz(n)), \bPhi^*(\bz(k))\right\rangle_{\HH}.
\end{align*}

Recall that in the complexification trick the associated function that maps the data to $\HH$ is given by $\hat\bPhi(\bz) = \Phi(\bx,\by) + i\Phi(\bx,\by) = \kappa_{\R}(\cdot,(\bx,\by)) + i\kappa_{\R}(\cdot,(\bx,\by))$, where $\Phi$ is the feature map of $\cH$ and $\kappa_{\R}$ its respective real kernel. Under this condition, the filter output at iteration $n$ takes the form
\begin{align}\label{EQ:output_c2}
\tilde d(n) = 4\mu \sum_{k=1}^{n-1} e(k)\cdot\kappa_{\R}\left((\bx(k),\by(k)), (\bx(n),\by(n)) \right).
\end{align}
This is exactly the same formula, which was obtained in the case of the complexified complex kernel LMS (except a rescaling). Thus, we deduce that, in this case, the standard complexified NCKLMS1 presented in \cite{Bouboulis_2011_10643} and the complexified augmented CKLMS are identical.

On the other hand, for the case of a pure complex kernel $\kappa_{\C}$, the filter output becomes
\begin{align}\label{EQ:output_c1}
\tilde d(n) = \mu\sum_{k=1}^{n-1} e(k) \kappa_{\C}(\bz(k), \bz(n)) + \mu\sum_{k=1}^{n-1} e(k)\kappa_{\C}^*(\bz(k), \bz(n)),
\end{align}
as $\left\langle\bPhi(\bz), \bPhi(\bc)\right\rangle_{\HH}=\kappa_{\C}(\bz,\bc)$, for any $\bz,\bc\in\C^\nu$.
In this case, it is evident that the augmented CKLMS (ACKLMS) will result to a different solution compared to that of NCKLMS2, as it exploits a richer representation. From (\ref{EQ:output_c1}), we deduce that the complexity of the normalised ACKLMS is of the same order as the complexity of NCKLMS2 and NCKLMS1.

\begin{table}
\begin{center}

\begin{tabular}{|c|c|}
\hline
\textbf{Name} & \textbf{Description} \\\hline
NCLMS & Normalized Complex LMS\\\hline
NACLMS & Normalized Augmented  Complex LMS\\
 & (Widely-Linear)\\\hline
NCKLMS1 & Normalized Complex Kernel LMS\\
 & (complexified real kernels) \\\hline
NCKLMS2 & Normalized Complex Kernel LMS\\
 & (pure complex kernels) \\\hline
NACKLMS & Normalized Augmented Complex Kernel LMS \\
 & (pure complex kernels)\\\hline
 CNGD & Complex Non-linear Gradient Descent \cite{Mandic_2009_10646}\\\hline
 MLP &  Multi Layer Perceptron \\
 & (50 nodes in the hidden layer)\\\hline
\end{tabular}
\caption{Acronyms of the Algorithms used in the paper.}\label{TAB:Acro}
\end{center}
\end{table}

\begin{figure*}
\begin{center}
\includegraphics[scale=0.45]{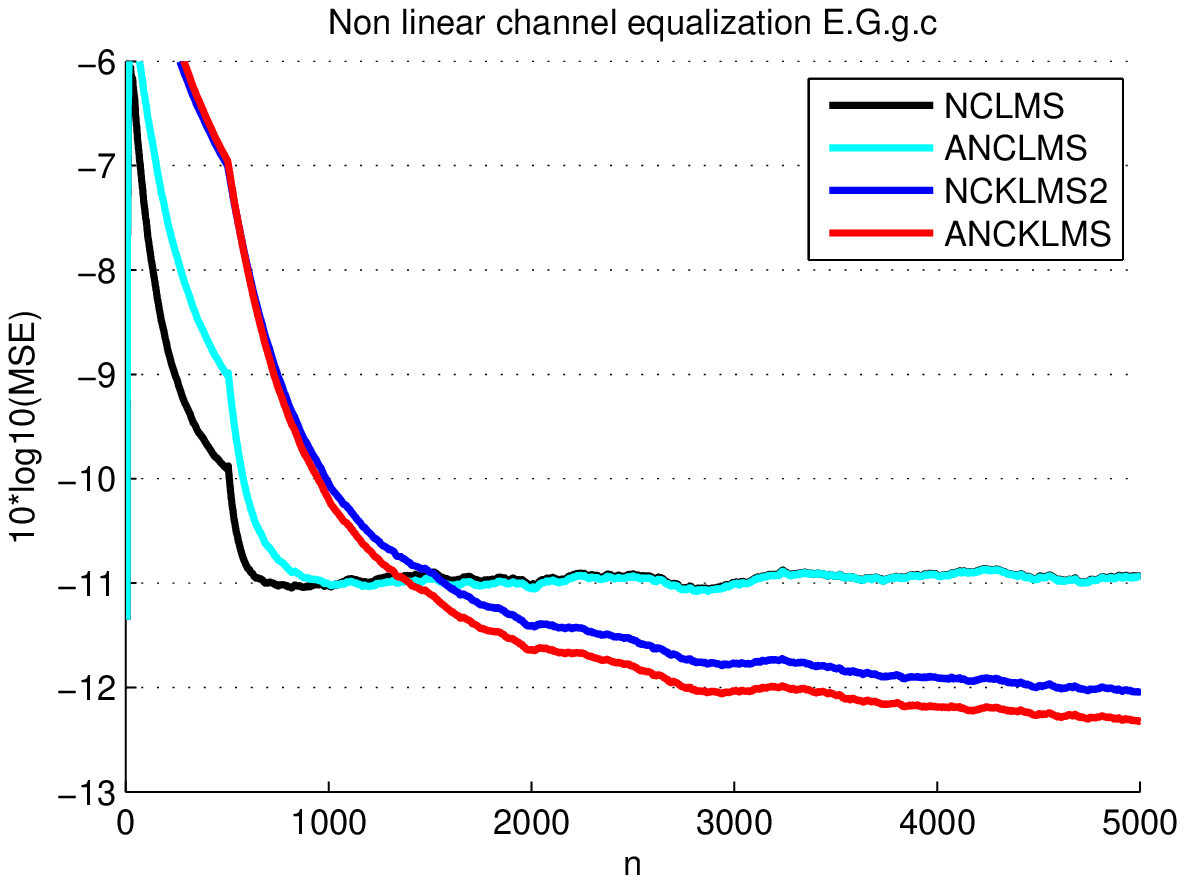}
\includegraphics[scale=0.45]{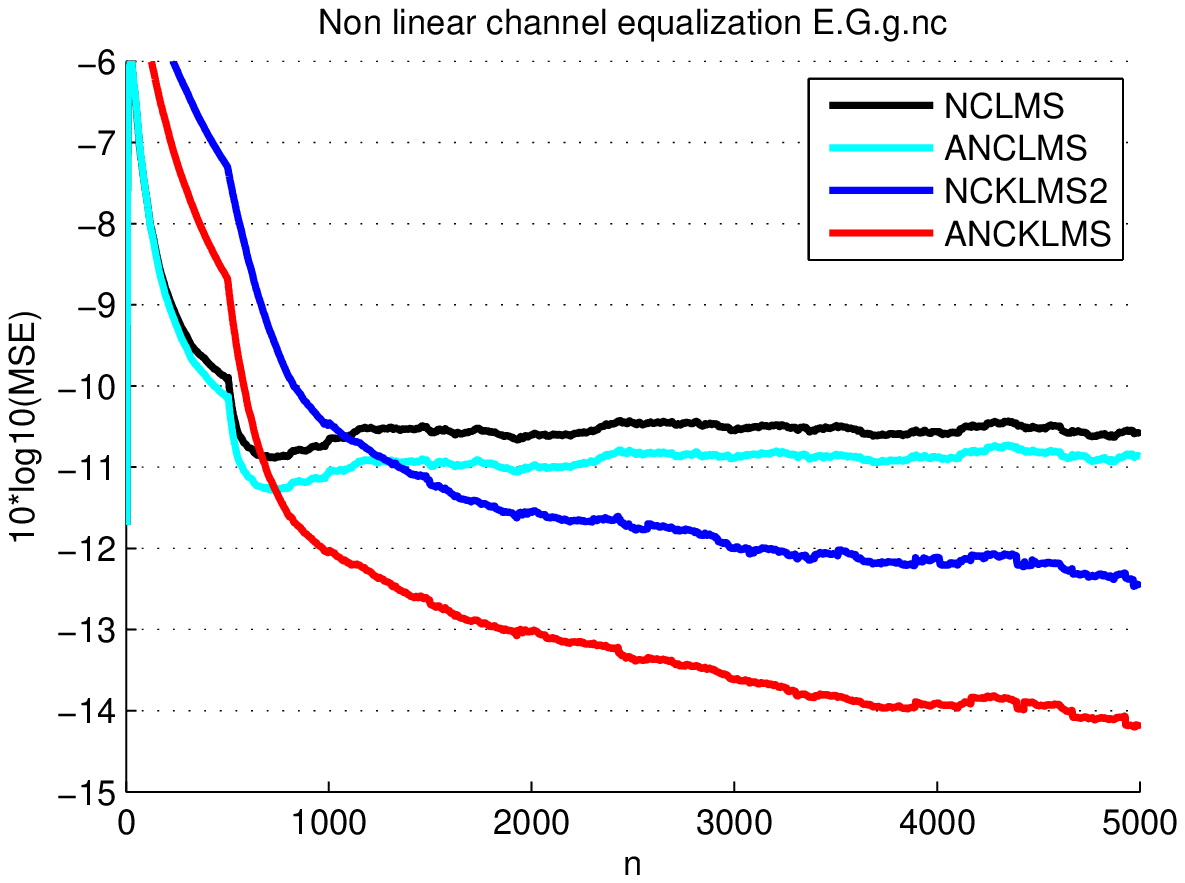}\\
(a)\hspace{17em} (b)
\end{center}
\caption{Learning curves for NCKLMS2 ($\mu=1/8$), NACKLMS, ($\mu=1/8$, $\sigma=10$), ÍCLMS ($\mu=1/16$) and widely linear ÍCLMS ($\mu=1/16$) (filter length $L=5$, delay $D=2$) for the soft nonlinear channel equalization problem, for (a) the circular input case, (b) the non-circular input case ($\rho=0.1$).}\label{FIG:equal1}
\end{figure*}

\section{Experiments}\label{SEC:EXPER}
The performance of the normalized augmented CKLMS (NACKLMS) has been tested in the context of a nonlinear channel equalization task. As in \cite{Bouboulis_2011_10643}, two nonlinear channels have been considered. The first channel (labeled as \textit{soft nonlinear channel} in the figures) consists of a linear filter: $t(n)= (-0.9+0.8i)\cdot s(n) + (0.6-0.7i)\cdot s(n-1)$
and a memoryless nonlinearity
$q(n) =\; t(n) + (0.1+0.15i)\cdot t^2(n) + (0.06+0.05i)\cdot t^3(n)$.
The second one (labeled as \textit{strong nonlinear channel} in the figures) comprises of the consists of the linear filter: $t(n)= (-0.9+0.8i)\cdot s(n) + (0.6-0.7i)\cdot s(n-1) + (-0.4 + 0.3i)\cdot s(n-2) +  (0.3 - 0.2i)\cdot s(n-3) + (-0.1i - 0.2i)\cdot s(n-4))$ and the nonlinearity:
$q(n) =\; t(n) + (0.2+0.25i)\cdot t^2(n) + (0.08+0.09i)\cdot t^3(n)$.
At the receiver end of the channels, the signal is corrupted by white Gaussian noise and then observed as $r(n)$. The level of the noise was set to 15dB. The input signal that was fed to the channels had the form
$s(n) = 0.70\left(\sqrt{1-\rho^2}X(n) + i\rho Y(n)\right)$,
where $X(n)$ and $Y(n)$ are gaussian random variables. This input is circular for $\rho=\sqrt{2}/2$ and highly non-circular if $\rho$ approaches 0 or 1 \cite{Adali_2010_10640}. To solve the channel equalization task, we apply the NACKLMS algorithm to the set of samples
$\left((r(n+D), r(n+D-1), \dots, r(n+D-L+1)), s(n)\right)$,
where $L>0$ is the filter length and $D$ the equalization time delay.

\begin{figure*}
\begin{center}
\includegraphics[scale=0.45]{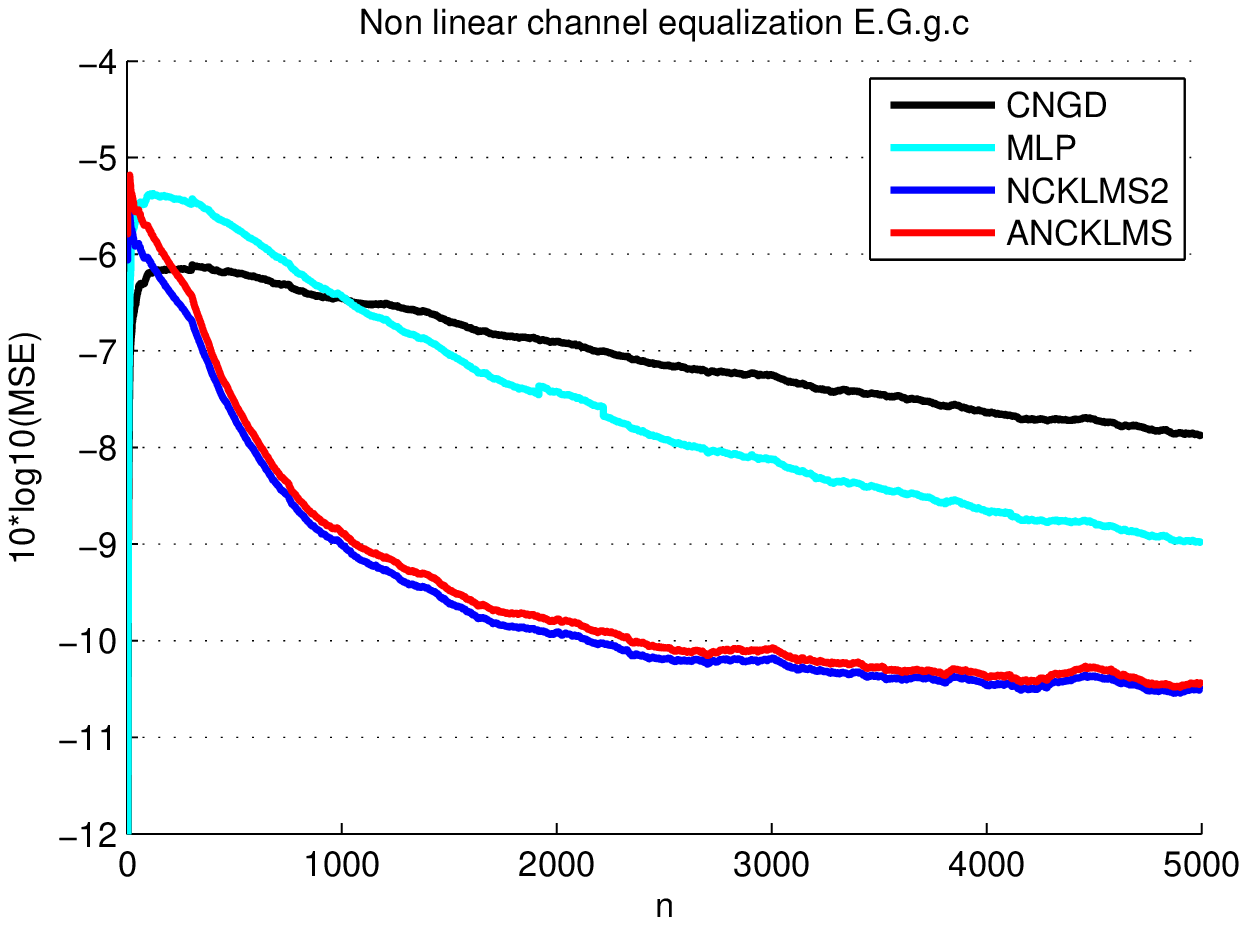}
\includegraphics[scale=0.45]{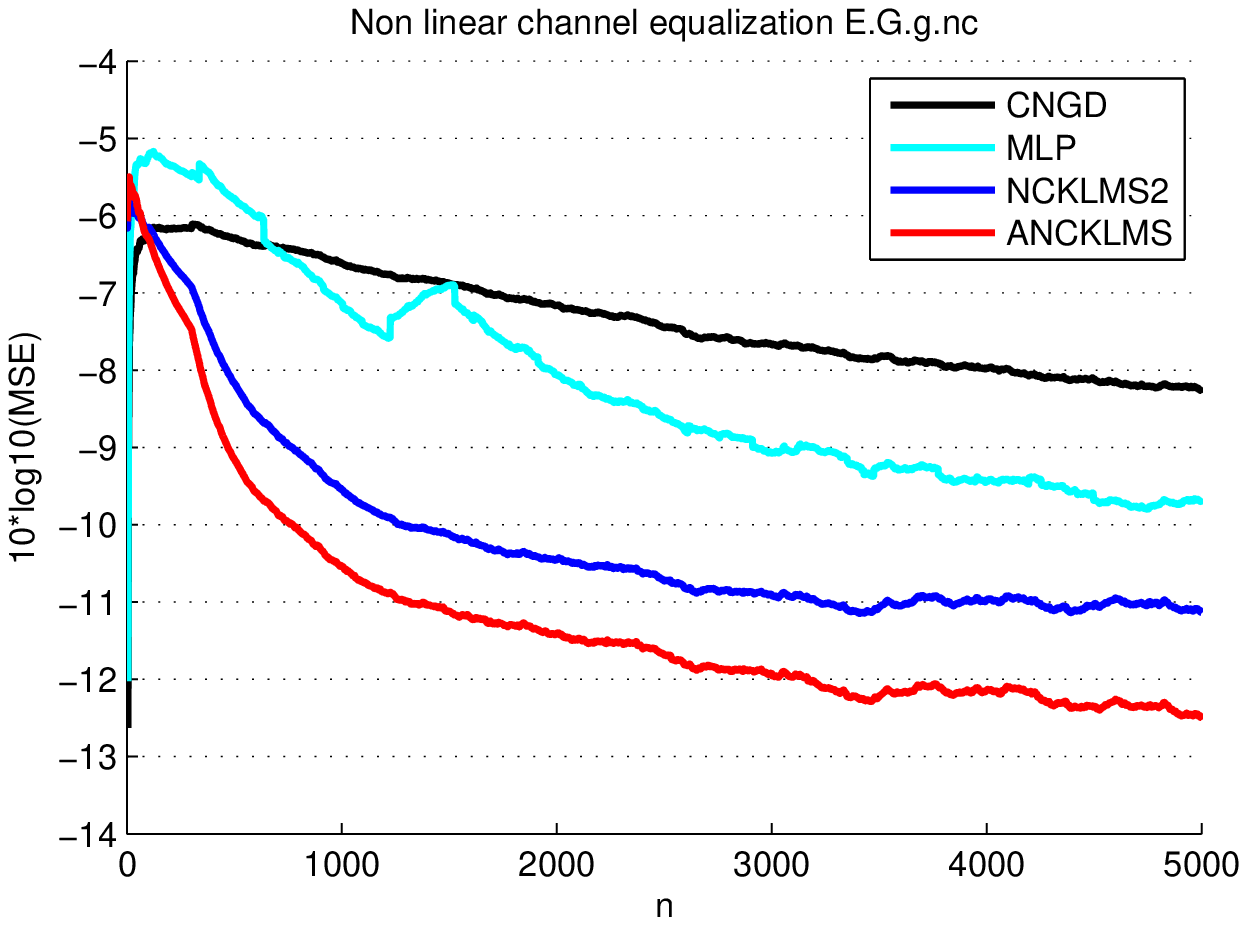}\\
(a)\hspace{17em} (b)
\end{center}
\caption{Learning curves for NCKLMS2 ($\mu=1/8$), NACKLMS, ($\mu=1/8$, $\sigma=15$), MLP ($\mu=0.0003$) and CNGD ($\mu=0.0005$) (filter length $L=5$, delay $D=2$) for the hard nonlinear channel equalization problem, for (a) the circular input case, (b) the non-circular input case ($\rho=0.1$).}\label{FIG:equal2}
\end{figure*}

Experiments were conducted on 100 sets of 5000 samples of the input signal considering both the circular and the non-circular cases. The results are compared with the NCLMS and the NACLMS (i.e., Normalized Augmented CLMS, or widely linear NCLMS as it is sometimes called) algorithms as well as two adaptive nonlinear algorithms: a) the CNGD algorithm \cite{Mandic_2009_10646} and a Multi Layer Perceptron (MLP) with 50 nodes in the hidden layer \cite{Adali_2010_10640}. In both cases, the complex $\tanh$ activation function was employed. Figure \ref{FIG:equal1} shows the learning curves of the pure complex kernel LMS (labeled as NCKLMS2 in the figures) \cite{Bouboulis_2011_10643} and the proposed NACKLMS,  using the complex Gaussian kernel (with $\sigma=10$), together with those obtained from the NCLMS and the WL-NCLMS algorithms.
Figure \ref{FIG:equal2}  shows the learning curves of NCKLMS2 and NACKLMS, using the complex Gaussian kernel (with $\sigma=15$), versus the CNGD and the $L$-50-1 MLP for the hard non-linear channel.

The novelty criterion (see \cite{Bouboulis_2011_10643}, \cite{Liu_2010_10644}) was used for the sparsification of the NCKLMS2 and NACKLMS  with $\delta_1=0.1$ and $\delta_2=0.2$. In both examples, NACKLMS considerably outperforms the linear, widely linear (i.e., NCLMS and WL-NCLMS) and nonlinear (CNGD and MLP) algorithms (see figures \ref{FIG:equal1}, \ref{FIG:equal2}). The NACKLMS also exhibits improved performance compared to the NCKLMS2 for non-circular input sources. Moreover, observe that while the gain of the NACLMS against NCLMS is rather rudimentary (smaller than 0.2 dB), the gains of NACKLMS over NCKLMS2 are significant (approximately 2dB). For circular signals, the two models (NCKLMS2 and NACKLMS) lead to almost identical results, as expected \cite{Picinbono_1994_10659, Picinbono_1995_10647}.

\section{Conclusions}\label{SEC:CONCL}
In this paper we proposed a method for widely linear estimation in complex RKHS, based on the framework presented in \cite{Bouboulis_2011_10643}. The developed rationale was applied to derive the augmented complex kernel LMS (NACKLMS).  Moreover, some important properties of widely linear estimation were discussed to provide a further understanding of the underlying toolbox. Experiments of the developed NACKLMS, for both circular and non-circular input data, showed a significant decrease in the steady state mean square error, compared with other known linear, widely linear and nonlinear techniques. Both theoretical and experimental results showed the importance of employing augmented filters versus linear ones in the complex domain.

% Can use something like this to put references on a page
% by themselves when using endfloat and the captionsoff option.
\ifCLASSOPTIONcaptionsoff
  \newpage
\fi

% trigger a \newpage just before the given reference
% number - used to balance the columns on the last page
% adjust value as needed - may need to be readjusted if
% the document is modified later
%\IEEEtriggeratref{8}
% The "triggered" command can be changed if desired:
%\IEEEtriggercmd{\enlargethispage{-5in}}

% references section

% can use a bibliography generated by BibTeX as a .bbl file
% BibTeX documentation can be easily obtained at:
% http://www.ctan.org/tex-archive/biblio/bibtex/contrib/doc/
% The IEEEtran BibTeX style support page is at:
% http://www.michaelshell.org/tex/ieeetran/bibtex/
\bibliographystyle{IEEEtran}
% argument is your BibTeX string definitions and bibliography database(s)
\bibliography{athensBIB}
%
% <OR> manually copy in the resultant .bbl file
% set second argument of \begin to the number of references
% (used to reserve space for the reference number labels box)

% biography section
%
% If you have an EPS/PDF photo (graphicx package needed) extra braces are
% needed around the contents of the optional argument to biography to prevent
% the LaTeX parser from getting confused when it sees the complicated
% \includegraphics command within an optional argument. (You could create
% your own custom macro containing the \includegraphics command to make things
% simpler here.)
%\begin{biography}[{\includegraphics[width=1in,height=1.25in,clip,keepaspectratio]{mshell}}]{Michael Shell}
% or if you just want to reserve a space for a photo:

% insert where needed to balance the two columns on the last page with
% biographies
%\newpage

% You can push biographies down or up by placing
% a \vfill before or after them. The appropriate
% use of \vfill depends on what kind of text is
% on the last page and whether or not the columns
% are being equalized.

%\vfill

% Can be used to pull up biographies so that the bottom of the last one
% is flush with the other column.
%\enlargethispage{-5in}

% that's all folks
\end{document}